\pdfoutput=1
\documentclass{article}

\usepackage[nonatbib,preprint]{neurips_2024}
\usepackage[numbers]{natbib}

\usepackage[utf8]{inputenc} 
\usepackage[T1]{fontenc}    
\usepackage{hyperref}       
\usepackage{url}            
\usepackage{booktabs}       
\usepackage{amsfonts}       
\usepackage{nicefrac}       
\usepackage{microtype}      
\usepackage{xcolor}         
\usepackage{amsmath}
\usepackage{amssymb}
\usepackage{mathtools}
\usepackage{amsthm}
\usepackage{listings}
\usepackage{bbm}

\usepackage{amsmath,amsfonts,bm}

\def\eqref#1{equation~\ref{#1}}

\def\1{\bm{1}}

\def\vc{{\bm{c}}}

\def\vx{{\bm{x}}}

\def\vz{{\bm{z}}}

\DeclareMathAlphabet{\mathsfit}{\encodingdefault}{\sfdefault}{m}{sl}
\SetMathAlphabet{\mathsfit}{bold}{\encodingdefault}{\sfdefault}{bx}{n}
\newcommand{\tens}[1]{\bm{\mathsfit{#1}}}

\def\tD{{\tens{D}}}

\newcommand{\R}{\mathbb{R}}

\DeclareMathOperator{\Tr}{Tr}

\DeclareMathOperator{\defeq}{\stackrel{\text{def}}{\; = \;}}
\DeclareMathOperator{\diag}{\text{diag}}

\theoremstyle{plain}
\newtheorem{theorem}{Theorem}[section]
\newtheorem{proposition}[theorem]{Proposition}

\theoremstyle{definition}
\newtheorem{definition}[theorem]{Definition}

\theoremstyle{remark}

\title{Towards an Improved Understanding and Utilization of Maximum Manifold Capacity Representations}

\author{%
  Rylan Schaeffer \thanks{Correspondence to \texttt{rschaef@cs.stanford.edu}} \\
  Stanford CS
\\
  \And
  Victor Lecomte$^{\dag}$ \\
  Stanford CS
\\
 \And
  Dhruv Pai\thanks{Equal contribution.} \\
  Stanford CS
\\
 \And
  Andres Carranza$^{\dag}$ \\
  Stanford CS
\\
 \And
  Berivan Isik$^{\dag}$ \\
  Stanford EE
\\
 \And
  Alyssa Unell$^{\dag}$ \\
  Stanford CS
\\
 \And
  Mikail Khona \\
  MIT Physics
\\
 \And
  Thomas Yerxa \\
  NYU Neural Science
\\
 \AND
  Yann LeCun \\
  NYU Data Science \& Meta AI FAIR
\\
 \And
  SueYeon Chung \\
  NYU Neural Science \& Flatiron Institute
\\
 \AND
  Andrey Gromov\thanks{Equal advising.} \\
  UMD Physics \& Meta AI FAIR
\\
 \And
  Ravid Shwartz-Ziv$^{\ddag}$ \\
  NYU Data Science
\\
 \And
  Sanmi Koyejo$^{\ddag}$ \\
  Stanford CS
}

\begin{document}

\maketitle

\begin{abstract}
Maximum Manifold Capacity Representations (MMCR) is a recent multi-view self-supervised learning (MVSSL) method that matches or surpasses other leading MVSSL methods.
MMCR is intriguing because it does not fit neatly into any of the commonplace MVSSL lineages, instead originating from a statistical mechanical perspective on the linear separability of data manifolds.
In this paper, we seek to improve our understanding and our utilization of MMCR. 
To better understand MMCR, we leverage tools from high dimensional probability to demonstrate that MMCR incentivizes alignment and uniformity of learned embeddings. We then leverage tools from information theory to show that such embeddings maximize a well-known lower bound on mutual information between views, thereby connecting the geometric perspective of MMCR to the information-theoretic perspective commonly discussed in MVSSL.
To better utilize MMCR, we mathematically predict and experimentally confirm non-monotonic changes in the pretraining loss akin to double descent but with respect to atypical hyperparameters. We also discover compute scaling laws that enable predicting the pretraining loss as a function of gradients steps, batch size, embedding dimension and number of views.
We then show that MMCR, originally applied to image data, is performant on multimodal image-text data.
By more deeply understanding the theoretical and empirical behavior of MMCR, our work reveals insights on improving MVSSL methods.
\end{abstract}

\section{Introduction}

Multi-View Self-Supervised Learning (MVSSL; also known as Joint-Embedding Self-Supervised Learning) is a powerful approach to unsupervised learning. The idea is to create multiple transformations, or ``views'', of unsupervised data, then use these views in a supervised-like manner to learn generally useful representations. MVSSL methods are diverse but can be loosely grouped into different families \citep{balestriero2023cookbook}: (1) contrastive, e.g., CPC \citep{oord2018representation}, MoCo 1 \citep{he2019moco},  SimCLR \citep{chen2020simple}, MoCo 2 \citep{chen2020mocov2}, CMC \citep{tian2020contrastive}, RPC \citep{tsai2021self} and TiCo \citep{zhu2022tico}; (2) clustering e.g., Noise-as-Targets \citep{bojanowski2017unsupervised}, DeepCluster \citep{caron2018deep}, Self-Labeling \citep{asano2019self}, Local Aggregation \citep{zhuang2019local}, SwAV \citep{caron2020unsupervised}; (3) distillation/momentum e.g., BYOL \citep{grill2020bootstrap}, DINO \citep{caron2021emerging}, SimSiam \citep{chen2021exploring}, TiCo \citep{zhu2022tico}; and (4) redundancy reduction e.g., Barlow Twins \citep{zbontar2021barlow}, VICReg \citep{bardes2021vicreg}, TiCo \citep{zhu2022tico}. Many MVSSL methods either explicitly originate from information theory \citep{oord2018representation, bachman2019learning} or can be understood from an information-theoretic perspective \citep{wang2020understanding,wu2020mutual, galvez2023role,shwartz2023information}.


\begin{figure*}
    \centering
    \includegraphics[width=\textwidth]{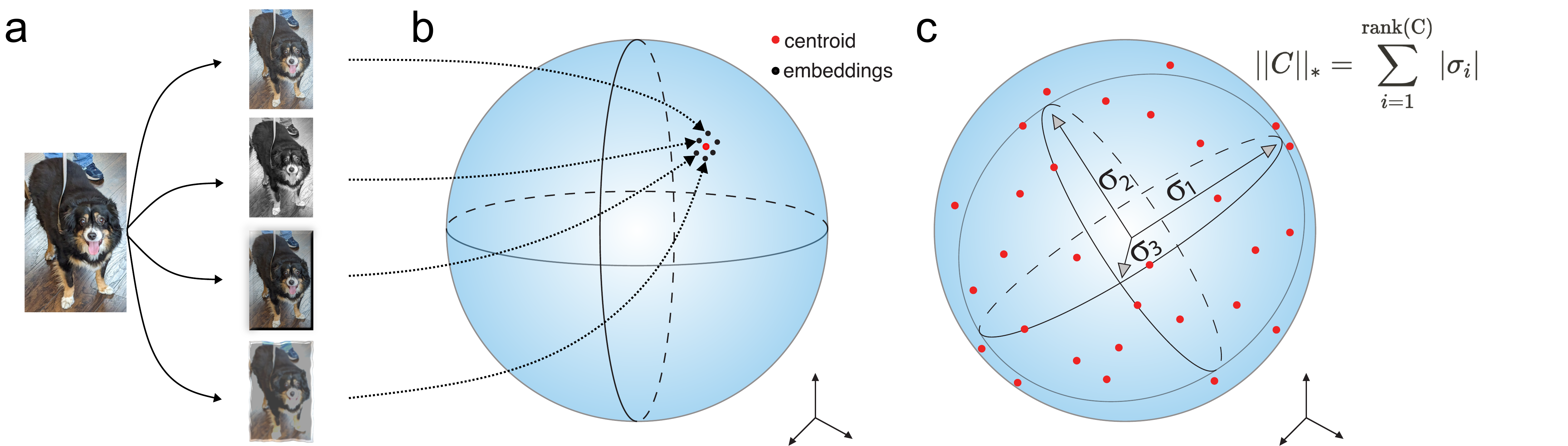}
    \caption{\textbf{Schematic of Maximum Manifold Capacity Representations (MMCR).} \textbf{(Left)} $K \geq 2$ views are generated of each datum, then embedded through a deep neural network on the surface of the hypersphere. Center: For each datum, the \emph{centroid} of the embeddings is computed. \textbf{(Right)} The MMCR pretraining loss, which is the negative nuclear norm of the centers, is then minimized.} 
    \label{fig:mmcr_schematic}
\end{figure*}

Recently, \citet{yerxa2023learning} proposed a new MVSSL method named Maximum Manifold Capacity Representations (MMCR) that achieves superior-to-similar performance compared with leading MVSSL methods.
MMCR is interesting for at least two reasons. 
Firstly, MMCR does not fit neatly into any of the MVSSL families: it is not (explicitly) contrastive, it performs no clustering, it leverages no distillation, and it does not (explicitly) reduce redundancy.
Secondly, unlike many MVSSL methods that originate in information theory, MMCR distances itself by pointing out that estimating mutual information in high dimensions has proven difficult and that more closely approximating mutual information may not improve representations; instead, MMCR's foundation lies in the statistical mechanical characterization of the linear separability of data manifolds.
In this work, we seek to better understand MMCR and utilize this understanding to drive implementation decisions. 
Our contributions are as follows:

\begin{enumerate}
    \item We leverage tools from high dimensional probability to show that embeddings with perfect invariance and perfect uniformity minimize the MMCR pretraining loss with high probability. This analysis involves bounding the MMCR pretraining loss, allowing us to define a ``pretraining percent error" for MMCR; this pretraining percent error then reveals two interesting empirical phenomena (below).
    \item We connect this distribution of embeddings to information theory by showing that such a distribution maximizes a well-known variational lower bound on the mutual information between embeddings of multiple views.
    \item Our analysis of the MMCR pretraining loss predicts a double descent-like behavior in the pretraining percent error as a function of two parameters: the number of manifolds $N$ and the embedding dimensionality $D$. We empirically test and confirm this prediction in ResNet-18s \cite{he2016deep} pretrained on STL-10 \cite{coates2011analysis}. This is notable because (to the best of our knowledge) double descent has not been observed in MVSSL and because these parameters differ from the typical double descent parameters (number of data and number of model parameters).
    \item Our pretraining percent error additionally enables comparing different hyperparameters on the MMCR pretraining loss -- an ability not commonly available in MVSSL methods -- which reveals the existence of compute scaling laws.
    \item We demonstrate that MMCR, originally proposed for images, can be similarly performant in the multi-modal image+text setting. We show that MMCR applied to image+text pairs can match CLIP \cite{radford2021learning} on DataComp Small containing 128M high quality image+caption pairs \cite{gadre2023datacomp} at smaller batch sizes but falls off with larger batch sizes.
\end{enumerate}

\section{Preliminaries}

\subsection{Multi-View Self-Supervised Learning (MVSSL)}

Let $f_{\theta}: \mathcal{X} \rightarrow \mathcal{Z}$ denote our neural network with parameters $\theta$. Suppose we have a dataset of $P$ points $\{ \vx_p \}_{p=1}^P$ and a set of random transformations (augmentations) $\mathcal{T}$ such as color jittering, Gaussian blur, solarization, etc..  For each datum $\vx_p$ in a batch of inputs, we sample $K$ transformations $t^{(1)}, t^{(2)}, ..., t^{(K)} \sim \mathcal{T}$ yielding a set of augmented views: $t^{(1)}(\vx_p), ..., t^{(K)}(\vx_p)$. We feed these transformed data into the network and obtain \emph{embeddings} $Z$:
\begin{align*}
    \vz_p^{(k)} \; \defeq \; f_{\theta}(t^{(k)}(\vx_p)) \in \mathcal{Z}.
\end{align*}

In practice, $\mathcal{Z}$ is commonly the $D$-dimensional hypersphere $\mathbb{S}^{D-1} \defeq \{ \vz \in \mathbb{R}^D : \vz^T \vz = 1 \}$ or $\mathbb{R}^D$. Given that we will later touch on information theory, we need notation to refer to the random variables; we use $Z_p^{(k)}$ to denote the random variable for the embedding whose realization is $\vz_p^{(k)}$.

\subsection{Maximum Manifold Capacity Representations} 

Maximum Manifold Capacity Representations (MMCR) \citep{yerxa2023learning} originates from classical results regarding performance of linear binary classifiers \citep{cover1965geometrical,gardner1987maximum, gardner1988space}.
Consider $P$ points (data) in $D$ dimensions, with arbitrarily assigned binary class labels; what is the probability that a linear binary classifier will be able to successfully classify the points?
Statistical mechanical calculations reveal that in the thermodynamic limit ($P, D\rightarrow \infty$; $P/D \rightarrow \alpha \in (0, \infty)$), a phase transition occurs at capacity $\alpha_c = 2$. More precisely, if $\alpha < \alpha_c$, the linear binary classifier will succeed with probability $1$; but if $\alpha > \alpha_c$, the classifier will succeed with probability $0$.
MMCR is based on an extension of this result from points to manifolds \citep{chung2018classification}. 
MMCR proceeds in the following manner: 
MMCR takes the embeddings output by the network and normalizes them to lie on the hypersphere: $\vz_p^{(1)}, ..., \vz_p^{(K)} \in \mathbb{S}^{D-1}$.
Then, MMCR computes the \emph{center} (average) of the embeddings for each datum:
\begin{equation}
    \vc_p \; \defeq \; \frac{1}{K} \sum_k \vz_p^{(k)}.
\end{equation}

Next, MMCR forms a $P \times D$ matrix $C$ where the $n$-th row of $C$ is the center $\vc_p$ and defines the loss:
\begin{equation}
\mathcal{L}_{MMCR} \; \defeq \; - \|C \|_* \; \defeq \; -\sum_{r=1}^{rank(C)} \; \sigma_r(C),    
\label{def:mmcr_loss}
\end{equation}
where $\sigma_r(C)$ is the $r$-th singular value of $C$ and $\|\cdot \|_*$ is the nuclear norm (trace norm, Schatten 1-norm).
Minimizing the MMCR loss means maximizing the nuclear norm of the mean matrix $C$.
Yerxa et. al (2023) \citep{yerxa2023learning} note that no closed form solution exists for singular values of an arbitrary matrix, but when $P=2, D=2$, a closed form solution exists that offers intuition: $\|C\|_*$ will be maximized when (i) the norm of each mean is maximized i.e., $\|\vc_p\|_2 = 1$ (recalling that $0 \leq \|\vc_p\| < 1$ since the embeddings live on the hypersphere), and (ii) the means $\vc_1, \vc_2$ are orthogonal to one another.
While we commend the authors for working to offer intuition, it is unclear to what extent the $P=2, D=2$ setting sheds light on MMCR in general, as MMCR was theoretically derived and numerically implemented in the large data and high dimension regime. We seek to generalize the dimensionality of this statement and examine the impact of MMCR across arbitrary manifolds and dimensions.

\section{A High-Dimensional Probability Analysis of Maximum Manifold Capacity Representations}

In this section, we derive and intuitively explain two properties of Maximum Manifold Capacity Representations (MMCR).
We specifically consider MMCR's regime of large number of patterns $P$ and high embedding dimension $D$. We contribute two results:

\begin{enumerate}
    \item The MMCR loss $\mathcal{L}_{MMCR}$ can be minimized by (a) making each center $\vc_p = \frac{1}{K}\sum_k \bm z_p^{(k)}$ lie on the surface of the hypersphere, and (b) making the distribution of centers as close to uniform on the hypersphere as possible.
    \item This configuration of centers maximizes a well-known variational lower bound on the mutual information between embeddings \citep{gallager1968information} that was recently used to study and unify several MVSSL families \citep{galvez2023role}.
\end{enumerate}

More formally, we begin by adopting two useful definitions from prior works \citep{wang2020understanding, galvez2023role}:\newline

\begin{definition}[Perfect Reconstruction]
We say a network $f_{\theta}$ achieves \textit{perfect reconstruction} if $\forall \vx \in \mathcal{X}, \forall \, t^{(1)}, t^{(2)} \in \mathcal{T}$, $\bm z^{(1)} = f_{\theta}(t^{(1)}(\vx)) = f_{\theta}(t^{(2)}(\vx)) = \bm z^{(2)}$.\newline
\label{def:perfect_reconstruction}
\end{definition}

\begin{definition}[Perfect Uniformity]
Let $p(Z)$ be the distribution over the network representations induced by the data sampling and transformation sampling distributions. We say a network $f_{\theta}$ achieves \textit{perfect uniformity} if the distribution $p(Z)$ is the uniform distribution on the hypersphere.\newline
\label{def:perfect_uniformity} 
\end{definition}

\begin{figure*}[t!]
    \centering
    \includegraphics[width=0.5\textwidth]{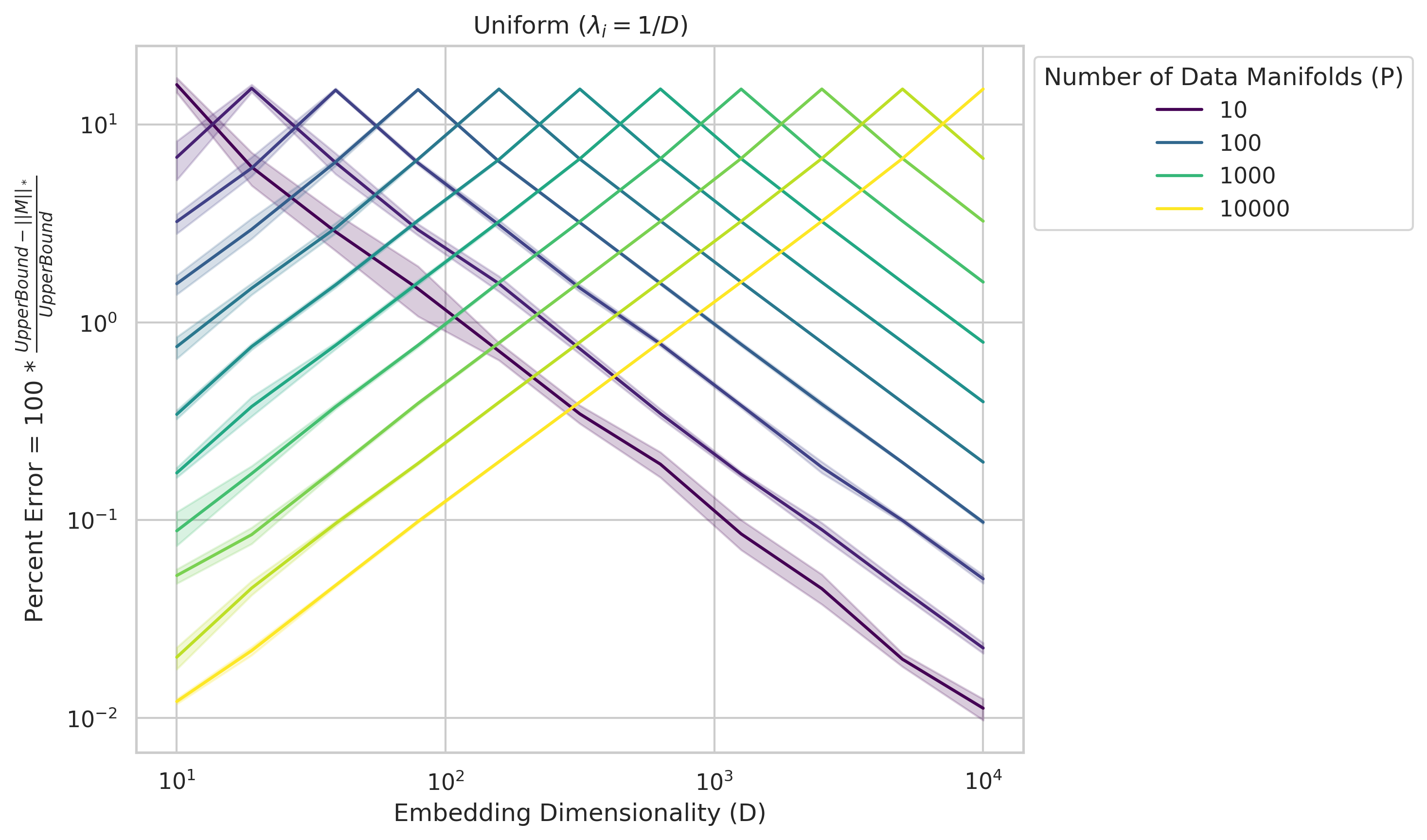}%
    \includegraphics[width=0.5\textwidth]{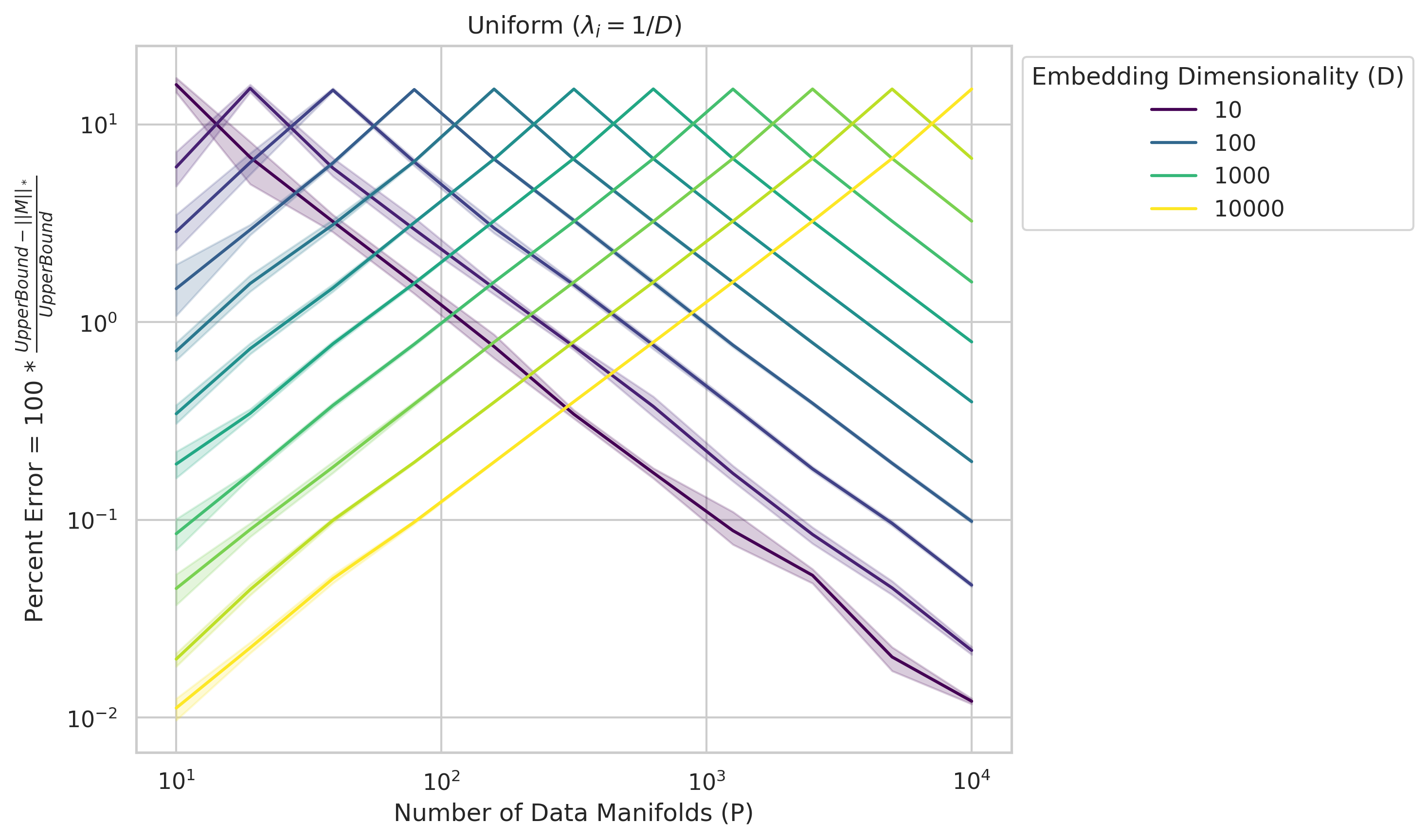}
    \caption{\textbf{Embeddings with perfect reconstruction and perfect uniformity achieve the lowest possible MMCR loss.} Away from the $P=D$ threshold, uniform random vectors achieve the theoretically derived upper bound on the nuclear norm of the mean matrix $M$ i.e. the lower bound on $\mathcal{L}_{MMCR}$. The gap between the network's loss and the lowest possible $\mathcal{L}_{MMCR}$ falls \textbf{(left)} $\propto P^{-1}$  or \textbf{(right)} $\propto D^{-1}$  away from the $P=D$ threshold.}
    \label{fig:numerical_simulations}
\end{figure*}

\looseness=-1
Intuitively, perfect reconstruction means that all views of the same datum are mapped by the network to the same embedding, and perfect uniformity means that the embeddings are distributed uniformly around the hypersphere. We will show that a network that achieves both perfect reconstruction and perfect uniformity obtains the lowest possible MMCR loss by first showing that $\mathcal{L}_{MMCR}$ has a lower bound and then showing that such a network achieves this bound.

\begin{proposition}
    Suppose that $\forall p \in [P], \vc_p^T \vc_p \leq 1$. Then,
    $0 \leq ||C||_* \leq \sqrt{P \, \min(P, D) }.$
    \label{prop:nuclear_norm_upper_bound}
\end{proposition}

\begin{proof}
    Let $\sigma_1, \ldots, \sigma_{\min(P,D)}$ denote the singular values of $C$, so that $\|C\|_* = \sum_{i=1}^{\min(P,D)} \sigma_i$. The lower bound follows by the fact that singular values are nonnegative. For the upper bound, we have
    \begin{align*}
        \sum_{i=1}^{\min(P,D)} \sigma_i^2 = \Tr\!\left[C C^T\right] = \sum_{n=1}^{P}\vc_p^T \vc_p \le P.
    \end{align*}
    Then, by Cauchy-Schwarz on the sequences $(1, \ldots, 1)$ and $\left(\sigma_1, \ldots, \sigma_{\min(P,D)}\right)$, we get
    \begin{align*}
        \sum_{i=1}^{\min(P,D)} \sigma_i
        &\le \sqrt{\left(\sum_{i=1}^{\min(P,D)} 1\right)\left(\sum_{i=1}^{\min(P,D)} \sigma_i^2\right)} \le \sqrt{\min(P,D) \, P}.
    \end{align*}
\end{proof}

\begin{proposition}
    Let $f_{\theta}$ achieve perfect reconstruction. Then, $\|\vc_p \|_2 = 1$ $\forall n$.
    \label{prop:means_on_sphere}
\end{proposition}

\begin{proof}
    Because $f_{\theta}$ achieves perfect reconstruction, $\forall n, \forall t^{(1)}, t^{(2)}$, $\bm z_p^{(1)} = \bm z_p^{(2)}$. Thus $\vc_p = (1/K)\sum_k \bm z_p^{(k)} = (1/K)\sum_k \bm z_p^{(1)} = \bm z_p^{(1)}$, and since $\|\bm z_p^{(1)}\|_2=1$, we have $\| \vc_p\|_2 = 1$.
\end{proof}

\begin{theorem}
    Let $f_{\theta}: \mathcal{X} \rightarrow \mathbb{S}^D$ be a network that achieves perfect reconstruction and perfect uniformity. Then $f_{\theta}$ achieves the lower bound of $\mathcal{L}_{MMCR}$ with high probability. Specifically:
    $$
    \|C\|_* = 
    \begin{cases}
        P(1-O(P/D)) & \text{if $P \le D$}\\
        \sqrt{PD}(1 - O(D/P)) & \text{if $P \ge D$}
    \end{cases},
    $$
    with high probability in $\min(P,D)$.
    \label{theorem:recon_plus_uniform_min_mmcr}
\end{theorem}

We defer the proof to Appendix~\ref{sec:proof_nuclear_norm_lower_bound} but offer intuition here.
To show the inequality in Proposition~\ref{prop:nuclear_norm_upper_bound} is roughly tight,
we need to show the singular values $\sigma_i$ are all roughly equal to each other.
When $P \ll D$, since $C$ has few rows $\vc_p$, they are almost perfectly orthogonal to each other, so all $P$ singular values will be $\approx \|\vc_p\| = 1$.
When $P \gg D$, since $C$ has many rows, for any $x \in \R^D$ the sum $\|Cx\|_2^2 = \sum_p (\vc_p^Tx)^2$ will be concentrated, so $C$ scales all vectors roughly equally, and therefore its $D$ singular values are all roughly equal to each other.
We confirm this via numerical simulations (Fig. \ref{fig:numerical_simulations}); for code, see Appendix \ref{app:python_code}. This demonstrates that the MMCR pretraining loss can be minimized by minimizing the distances of all embeddings corresponding to the same datum and maximizing the distances of all data's centers.

\section{An Information Theoretic Understanding of Maximum Manifold Capacity Representations}

Many MVSSL methods originate in information theory or can be understood from an information theoretic perspective \cite{oord2018representation, bachman2019learning,wang2020understanding,wu2020mutual, galvez2023role,shwartz2023information}. Based on our newfound understanding of what distributions of embeddings MMCR incentivizes, how can we connect MMCR's statistical mechanical geometric viewpoint to an information theoretic viewpoint? The answer is that MMCR incentivizes representations that maximize a lower bound on the mutual information shared by two embeddings corresponding to two views of the same datum.

Consider the mutual information between the embeddings of two different views $Z^{(1)}$ and $Z^{(2)}$ of some input datum. The mutual information between the two views must be at least as great as the sum of two terms: the ability of one embedding to ``reconstruct'' the other, plus the entropy of the embeddings \cite{gallager1968information}:
\begin{equation}
    \begin{aligned}
    I[Z^{(1)}; Z^{(2)}] \geq\underbrace{\mathbb{E}_{p(Z^{(1)}, Z^{(2)})}[\log q(Z^{(1)} | Z^{(2)})]}_{\text{Reconstruction}} + \underbrace{H[Z^{(1)}]}_{\text{Entropy}},
    \end{aligned}
\label{eq:mutual_information_lower_bound}
\end{equation}
\noindent where $q(Z^{(1)} | Z^{(2)})$ is a variational distribution because the true distribution $p(Z^{(1)} | Z^{(2)})$ is unknown. This bound is well-known, e.g., \cite{cover1965geometrical,wang2020understanding, galvez2023role}, but we repeat them to show how MMCR connects to an information-theoretic perspective.

\begin{proposition}
    For any distribution on a bounded space, the uniform distribution has maximum entropy.
\end{proposition}

\begin{theorem}
Let $f_{\theta}: \mathcal{X} \rightarrow \mathbb{S}^D$ be a network, and let the number of views per datum $K$ be a constant. Let $\mathcal{Q}$ be the variational family of distributions on the hypersphere. Then $f_{\theta}$ maximizes the mutual information lower bound (Eqn. \ref{eq:mutual_information_lower_bound}) if and only if $f_{\theta}$ achieves perfect reconstruction and perfect uniformity. \label{theorem:recon_plus_uniform_max_mi_lower_bound}
\end{theorem}

\begin{proof}
    Perfect reconstruction maximizes reconstruction term. Perfect uniformity maximizes entropy.
\end{proof}

Based on this result, we can now understand that a minimizer of the MMCR pretraining loss is a maximizer of the lower bound of the mutual information between two embeddings of two transformations of the same datum.

\begin{theorem}
    Let $f_{\theta^*}$ be a network that achieves perfect reconstruction and perfect uniformity, let the number of views per datum $K$ be a constant, and let $\mathcal{Q}$ be the variational family of distributions on the hypersphere. Then $f_{\theta^*}$ is both a minimizer of $\mathcal{L}_{MMCR}$ and a maximizer of the variational lower bound of mutual information Eqn. \ref{eq:mutual_information_lower_bound}.
\end{theorem}
\begin{proof}
    The proof follows from Theorem \ref{theorem:recon_plus_uniform_min_mmcr} and Theorem \ref{theorem:recon_plus_uniform_max_mi_lower_bound}.
\end{proof}

\section{Double Descent in Maximum Manifold Capacity Representations Pretraining Loss}

An unexpected and interesting insight from our high-dimensional probability analysis (Theorem \ref{theorem:recon_plus_uniform_min_mmcr}) is a prediction that the Maximum Manifold Capacity Representations (MMCR) pretraining loss should also exhibit a non-monotonic double descent-like behavior in its pretraining loss. Double descent is a well-known machine learning phenomenon where the test loss exhibits non-monotonic changes as a function of the total number of data and the number of model parameters \cite{vallet1989hebb,krogh1991simple,geman1992neural,krogh1992generalization,opper1995statistical,duin2000classifiers, spigler2018jamming, belkin2019reconciling, bartlett2020benign, belkin2020twomodels, nakkiran2021deep, poggio2019double, advani2020high, liang2020just, adlam2020understanding, rocks2022memorizing, rocks2021geometry, rocks2022bias, mei2022generalization, hastie2022surprises, bach2023highdimensional,schaeffer2023double, curth2023u,schaeffer2023divergence,schaeffer2024doubledescentdemystified}. However, our analysis suggests that this double descent-like behavior should occur with respect to atypical parameters: the number of manifolds $P$ and the number of dimensions $D$, rather than the number of data and the number of model parameters. Specifically, our theory predicts that the highest pretraining error should occur exactly at the threshold $P = D$, with pretraining error falling on either side of the threshold. Indeed, we discovered that Yerxa et al. (2023) \citep{yerxa2023learning} contains preliminary evidence of such a phenomenon (App Fig. 8); however, the double descent-like effect in the figure was neither commented on nor explored further, and the swept hyperparameters were suggestive but insufficient to be conclusive\footnote{Specifically, the purpose of the figure from Yerxa et. al (2023) \citep{yerxa2023learning} was to study the effect of number of manifolds, meaning that the data included only a single embedding dimension ($D=512$) and only one number of manifolds to the left of the threshold ($P=256$).}.

\begin{figure*}[t!]
    \centering
    \includegraphics[width=\textwidth]{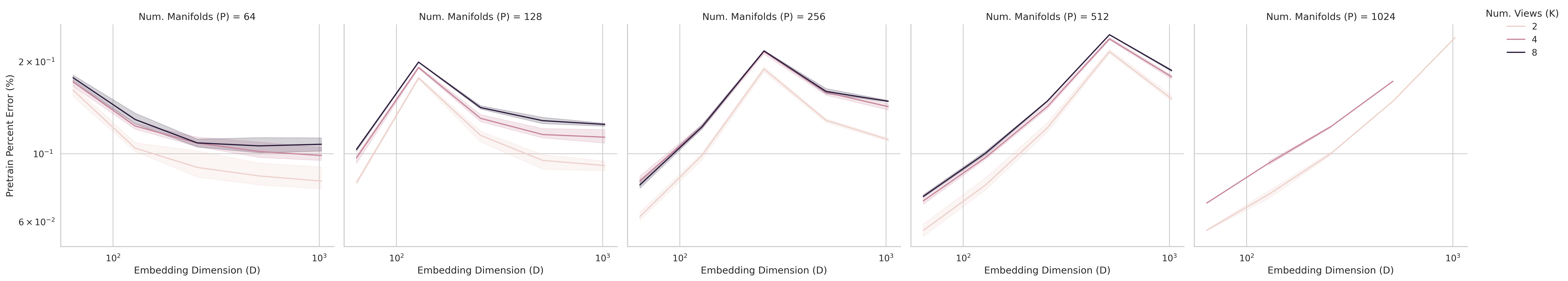}
    \includegraphics[width=\textwidth]{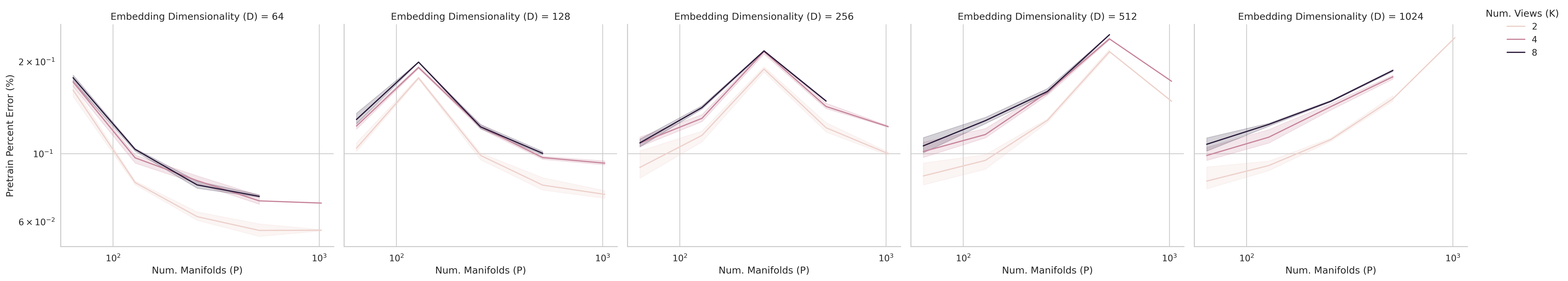}
    \caption{\textbf{Double-Descent in Maximum Manifold Capacity Representations.} As predicted mathematically, MMCR's pretraining percent error $\defeq (\sqrt{P \min(P, D)} - ||C||_*) / \sqrt{P \min(P, D)}$ exhibits non-monotonic double descent-like behavior, peaking when the number of data points $P$ equals the number of dimensions $D$. On either side of the $P=D$ threshold, the pretraining percent error falls. Networks are ResNet-18s pretrained on STL-10's ``unlabeled" split.}
    \label{fig:double_descent}
\end{figure*}

To compare losses across different pairs of hyperparameters number of points $P$ and data dimensionality $D$, we use our MMCR pretraining bound (Prop. \ref{prop:nuclear_norm_upper_bound}) to define a pretraining percent error:
\begin{equation}
    \text{Pretraining Percent Error}(C) \defeq \frac{\sqrt{P \min(P, D)} - ||C||_*}{ \sqrt{P \min(P, D)}}.
\end{equation}

To experimentally test our prediction, we pretrained ResNet-18 convolutional neural networks \cite{he2016deep} on STL-10 \cite{coates2011analysis}, a dataset similar to CIFAR-10 but higher resolution (96x96x3) and containing an additional unlabeled split of 100000 images. We swept $P \in \{64, 128, 256, 512, 1024\} \times D \in \{64, 128, 256, 512, 1024\} \times K \in \{2, 4, 8\}$, where $K$ is the number of views. 
For all combinations of number of points $P$, number of dimensions $D$ and number of views $K$, we found that the pretraining percent error peaked when $P=D$ (Fig. \ref{fig:double_descent}). On either side of the $P=D$ threshold, the pretraining percent error declined. Earlier work found that self-supervised learning doesn't seem to produce double descent like behavior \citep{lupidi2023does}, albeit for autoencoders. To the best of our knowledge, this is the first observation of double descent-like behavior in Multi-View Self-Supervised Learning, especially as a function of two unusual quantities (the number of data points $P$ and number of embedding dimensions $D$) rather than the classical double descent quantities (total number of data points and number of model parameters).

\begin{figure*}[t!]
    \centering
    \includegraphics[width=0.9\textwidth]{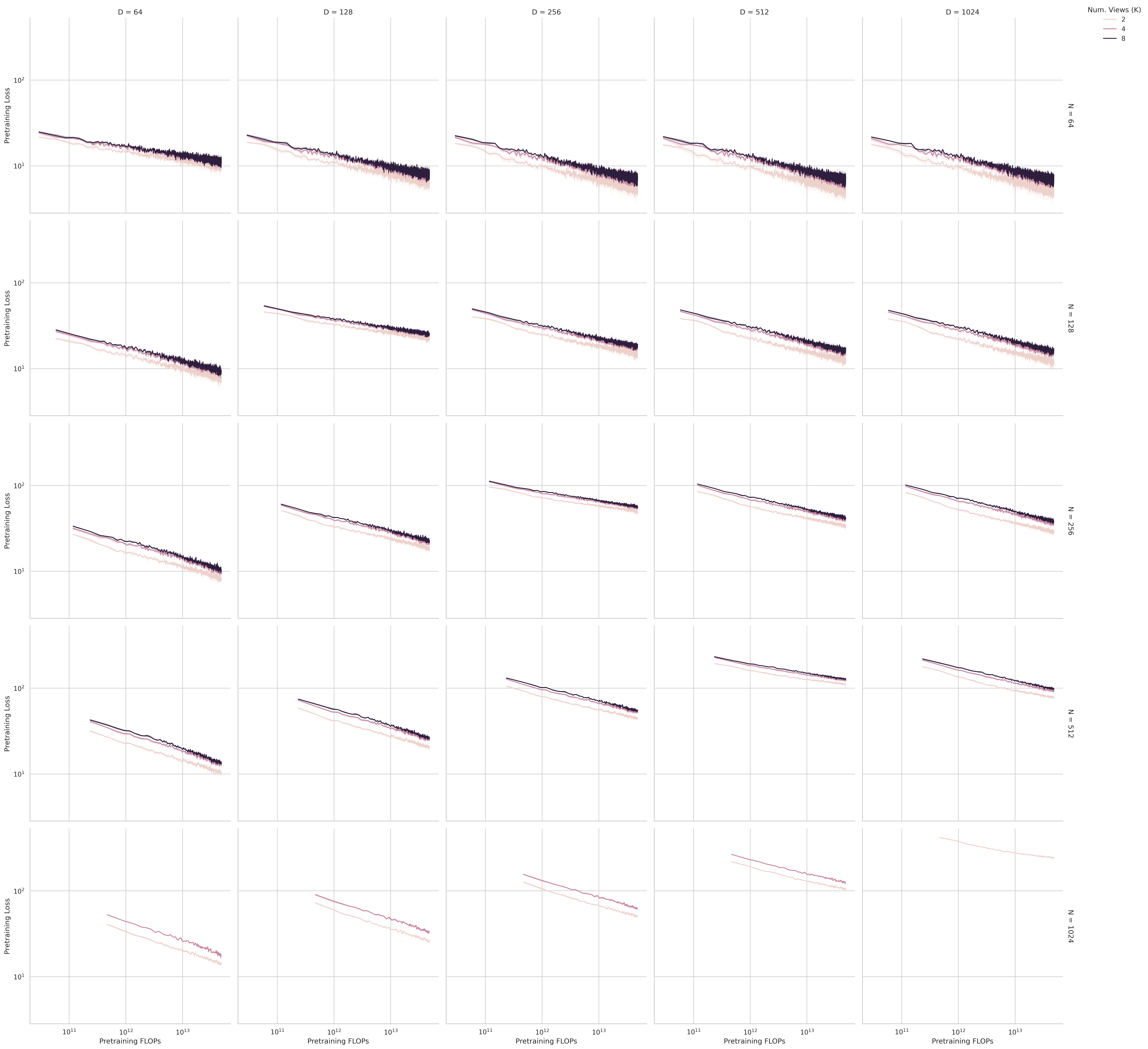}
    \caption{\textbf{Compute Scaling Laws.} For all values of number of points $P$ (equivalently, batch size), embedding dimension $D$ and number of views per datum $K$, the pretraining percent error falls predictably as a power law with the amount of compute i.e. total floating point operations. Consistent with the double descent-like findings in Fig. \ref{fig:double_descent}, the on-diagonal subfigures (corresponding to $P = D$) exhibit higher initial pretraining percent errors and less steep slopes with compute than the off-diagonal subfigures (corresponding to $P \neq D$).}
    \label{fig:scaling_laws}
\end{figure*}

\section{Compute Scaling Laws in Maximum Manifold Capacity Representations}

In many MVSSL methods, changing hyperparameters often renders the pretraining losses incommensurate, making comparisons between runs difficult if not impossible. However, because the MMCR pretraining percent error yields a quantity bounded between $0$ and $1$, we can compare different training runs with different hyperparameter values for the number of data points $P$ and data dimensionality $D$. Performing such a comparison yields a second interesting empirical phenomenon: compute neural scaling laws in the MMCR pretraining percent error. Scaling laws are another wide-spread phenomenon of interest in machine learning where the pretraining loss follows a predictable power law-like trend with respect to specific quantities such as number of parameters, number of data or amount of compute \cite{hestness2017deep,rosenfeld2019constructive,henighan2020scaling,kaplan2020scaling,gordon2021data,hernandez2021scaling,jones2021scaling,zhai2022scaling,hoffmann2022training,clark2022unified,neumann2022scaling,hernandez2022scaling,maloney2022solvable,schaeffer2023emergent,sardana2023chinchillaoptimal,muennighoff2024scaling,besiroglu2024chinchilla,gadre2024language, schaeffer2024predicting}.

By plotting the ResNet-18 networks pretrained on STL-10, one can clearly see power law scaling in the pretraining percent error with the amount of compute for all number of points $P$, embedding dimensions $D$, and number of views $K$ (Fig. \ref{fig:scaling_laws}). A key detail is that these neural scaling curves highlight the double descent-like behavior: the on-diagonal subfigures (corresponding to runs where $P=D$) have both higher pretraining percent error and a less steep slope for the pretraining percent error, meaning that the pretraining percent error starts higher and falls more slowly. The takeway is that practictioners would be well advised to not pretrain networks where the number of points $P$ (i.e. the batch size) equals the embedding dimension $D$.

\section{Multi-Modality in Maximum Manifold Capacity Representations}

\begin{figure*}
    \centering
    \includegraphics[width=0.49\columnwidth]{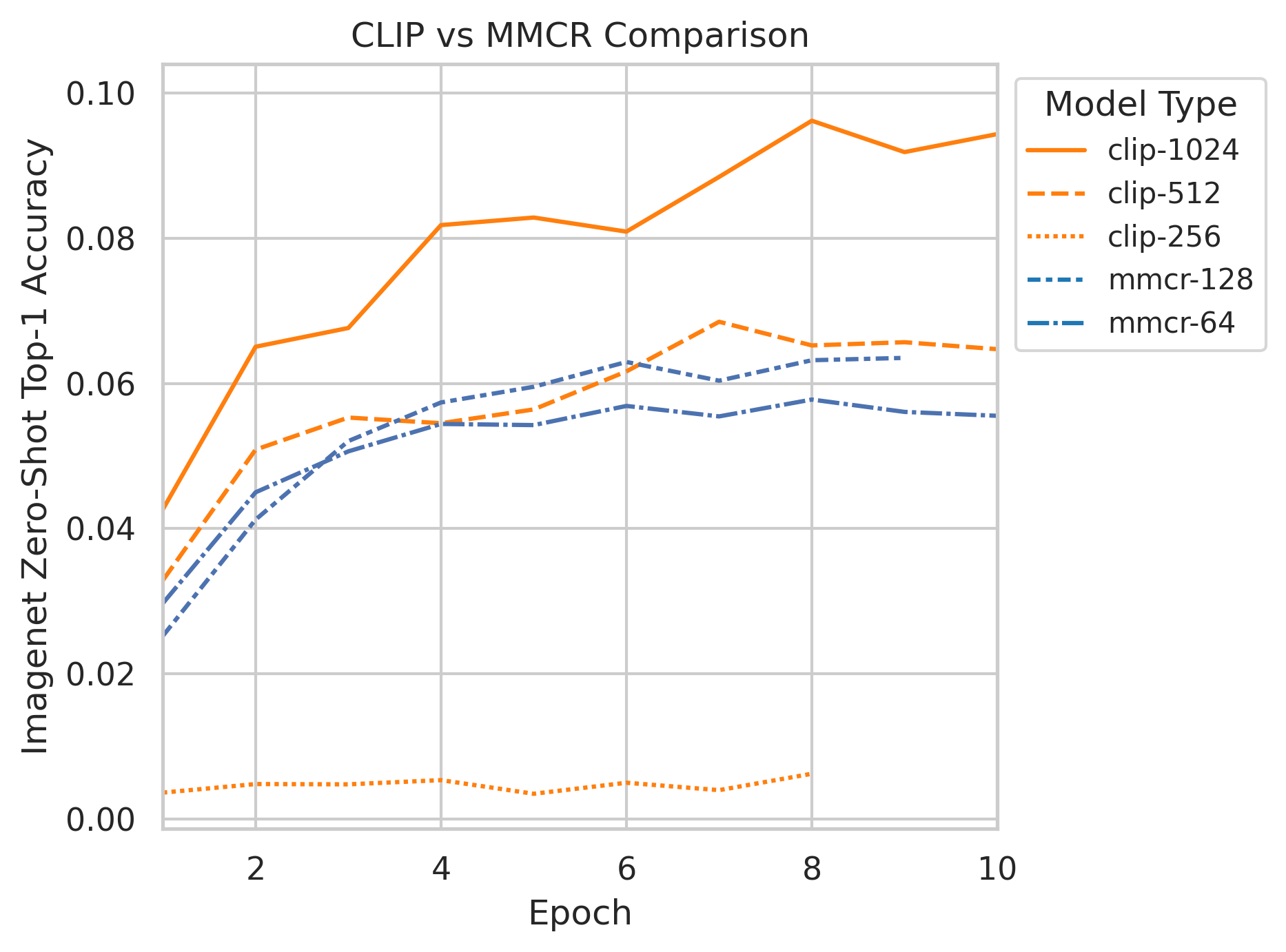}%
    \includegraphics[width=0.49\columnwidth]{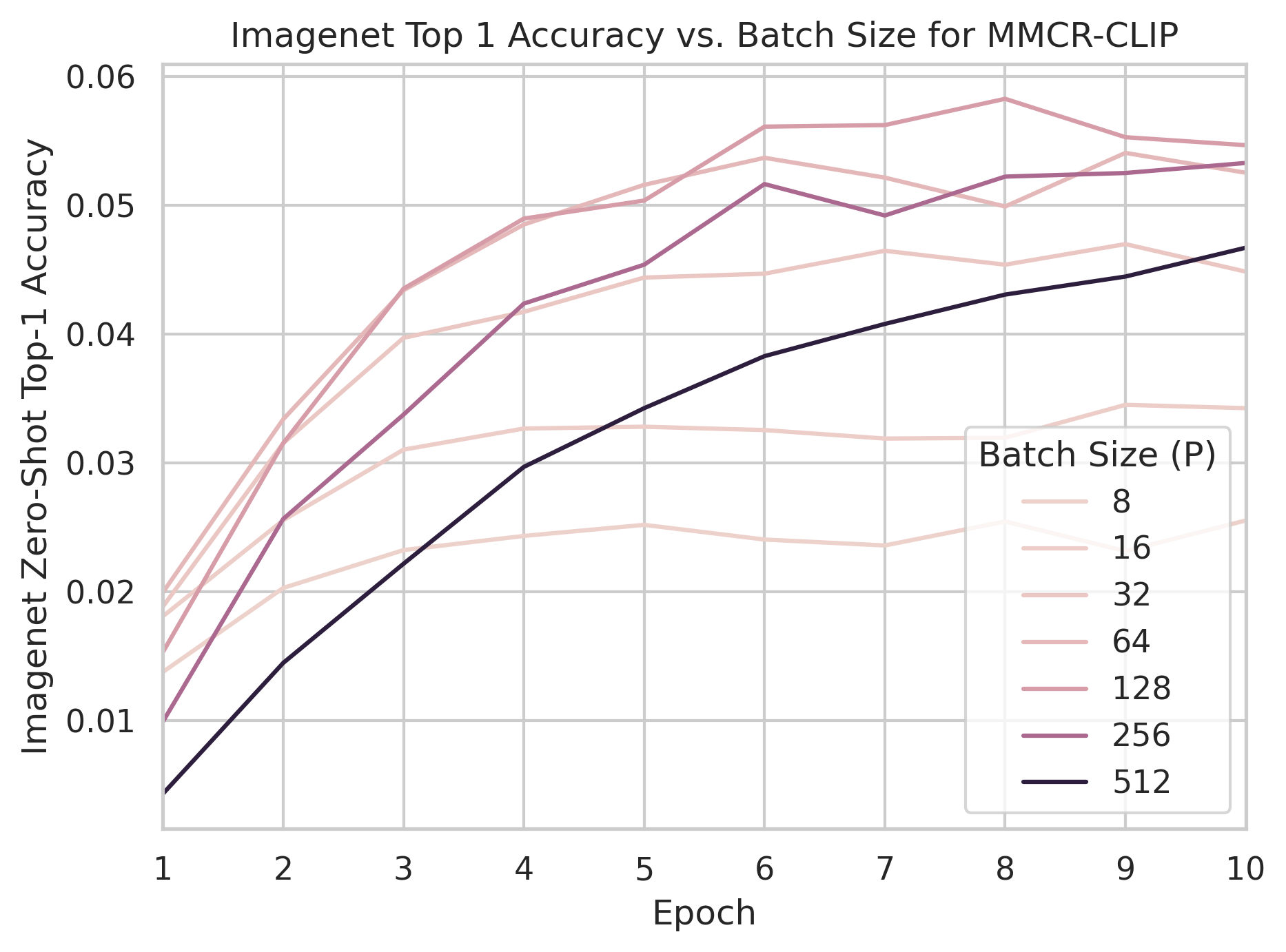}
     \caption{\textbf{Multimodal MMCR on Image-Text Caption Pairs.} Left: Multimodal MMCR vs Contrastive Language-Image Pretraining (CLIP) performance on ImageNet measured in zero-shot top-1 accuracy. Multimodal MMCR outperforms CLIP for smaller batch sizes but underperforms CLIP for larger batch sizes. Right: Imagenet top-1 accuracy sweep over batch sizes for MMCR. Unlike CLIP, MMCR exhibits non-monotonic performance scaling with batch size, and best results are found at intermediate batch sizes. To generate strong validation performance scaling behavior, MMCR requires that both batch size and dimension increase simultaneously.} 
    \label{fig:mmcr_vs_clip}
\end{figure*}

We next demonstrate that MMCR can be high-performing in a decidedly more challenging setting: multimodal self-supervised learning. Specifically, we consider the setting of OpenAI's Contrastive Language-Image Pretraining model (CLIP) \cite{radford2021learning}, in which two different networks are pretrained on image-text caption pairs.
In this multimodal setting, two networks $f_\theta$ and $g_{\theta'}$ embed data from two different data domains $X$ and $Y$. $X$ and $Y$ are paired, such that every example in $X$ has a corresponding positive pair in $Y$ and vice versa. As such, from an MMCR perspective, $X$ and $Y$ can be understood as two "views" of the same underlying object. The optimal transformed embeddings $f_\theta(X)$ and $g_{\theta'}(Y)$ therefore should map to the same space, and we can use our improved understanding of MMCR to train these optimal networks.
The notable difference between this setting and the commonplace MVSSL setting is first that $X$ and $Y$ might represent extremely different distributions in practice, and second $f_\theta$ and $g_\theta$ are two separate and different neural network architectures. CLIP is a prominent example of such a cross-modal feature alignment task between a text encoder and an image encoder \cite{radford2021learning}. In this paper, we investigate whether applying the MMCR objective to the CLIP setting can improve the quality of learned representations.

In image-text alignment, we have access to image-text pairs, which are respectively fed through a vision encoder (here, a ResNet-50) and a text encoder (a transformer \cite{vaswani2017attention}). We apply the MMCR objective between the embeddings produced by the two modalities.
We base our Multimodal MMCR experiments off of the open-source CLIP implementation OpenCLIP \cite{cherti2022reproducible}. We apply Multimodal MMCR to DataComp-Small and compare zero-shot Imagenet performance with the standard CLIP objective, which is equivalent to SimCLR with $\tau = 1$. DataComp-Small is the smallest version of the curated DataComp dataset family for training CLIP-style models \cite{gadre2023datacomp}. This dataset consists of 128 million high-quality image and text pairs that can be used in multimodal training.

We found that convergence of MMCR in the image-text mapping setting is highly dependent on learning rate, and models will fail to converge for learning rates above $\approx 1e-4$. For all runs, we set our Multimodal MMCR learning rate as $1e-4$ and our normal CLIP learning rate as $1e-3$. With the standard CLIP embedding size of $D=1024$, we swept performance of our models over the critical hyperparameter of batch size ($N$), finding the optimal batch size to be 128. We compare the performance of the optimal batch size Multimodal MMCR to normal CLIP (Fig. \ref{fig:mmcr_vs_clip}). We find that while Multimodal MMCR outperforms CLIP at small batch sizes ($<$ 512) and remains competitive with CLIP with a batch size of 512, it underperforms CLIP at higher batch sizes. The CLIP loss is a batch contrastive method, and thus benefits directly from increasing batch size. MMCR, however, is simultaneously batch and dimension contrastive, and as a result to achieve similar scaling it is likely Multimodal MMCR would need to increase the size of its latent embedding space beyond $1024$ \cite{garrido2023duality}.

\section{Relationship of MMCR to the Duality of Sample-Contrastive and Dimension-Contrastive Self-Supervised Learning}

In their ICLR 2023 paper ``On the Duality Between Contrastive and Non-Contrastive Self-Supervised Learning", Garrido et. al (2023) \citep{garrido2023duality} noted that contrastive (also known as sample-contrastive) and non-contrastive (also known as dimension-contrastive) SSL methods can be seen as two sides of the same coin. Specifically, letting $Z \in \mathbb{R}^{PK \times D}$ denote the matrix of stacked embeddings, then sample-contrastive methods (e.g., SimCLR) incentivize entropy via:
\begin{equation*}
    \mathcal{L}_{\text{Sample-Contrastive}} \defeq ||Z Z^T - \diag(Z Z^T)||_F^2,
    \label{eqn:sample_contrastive_loss}
\end{equation*}

whereas dimension-contrastive methods (e.g., BarlowTwins) incentivize entropy via:
\begin{equation*}
    \mathcal{L}_{\text{Dimension-Contrastive}} \defeq ||Z^T Z - \diag(Z^T Z)||_F^2.
    \label{eqn:dim_contrastive_loss}
\end{equation*}

Both families also include an invariant loss $\mathcal{L}_{\text{Invariance}}$ as part of the total loss, typically the MSE between the positive pairs. We observe that both families of loss aim to maximize on-diagonal elements through their invariance loss and minimize off-diagonal elements through their contrastive losses, on a batch-wise or dimension-wise correlation matrix respectively. In both cases, the loss functions aim to maximize the spectra of their matrices (either $ZZ^T$ or $Z^T Z$), and given that these matrices have related spectra, one might wonder why not maximize the spectrum of $Z$ directly? Maximizing $Z$'s spectrum is qualitatively what MMCR aims to do via its nuclear norm-based loss.

\section{Discussion}

In this paper, we theoretically and empirically studied Maximum Manifold Capacity Representations (MMCR) \cite{yerxa2023learning}, a recent high-performing multi-view self-supervised learning (MVSSL) method that has demonstrated competitive-to-superior performance when compared to established methods in the field.  Our investigation into MMCR has two primary objectives: enhancing theoretical comprehension and expanding practical usage.

Our theoretical exploration of MMCR reveals its potential to be understood not only through its original geometric lens but also from an information-theoretic standpoint. The nuclear norm-based objective of MMCR correlates with the maximization of a recognized mutual information lower bound between views. This insight provides a deeper understanding of MMCR's operational mechanics. Additionally, we identified a non-monotonic behavior in pretraining loss associated with MMCR, reminiscent of the double descent phenomenon, but influenced by non-traditional hyperparameters. Furthermore, the compute scaling curves, derived from MMCR's loss function, facilitate a direct comparison of hyperparameters, marking an innovative step forward in hyperparameter evaluation and selection within the MVSSL framework.

Expanding MMCR's application beyond its initial success in image data, we demonstrate efficacy in multimodal image-text scenarios. Leveraging our refined understanding of MMCR's hyperparameters, we illustrate its adaptability and robustness across diverse data types. This extension underscores MMCR's versatility, attributed to its geometric foundations, and opens up new avenues for its application in various MVSSL challenges.

In conclusion, our deeper theoretical insights into MMCR not only elucidate its underlying principles but also enable the development of more versatile and effective implementations. By harnessing these theoretical advancements, MMCR's applicability across a broader spectrum of MVSSL problems is significantly enhanced, promising new directions for future research in the field.

Note: This manuscript appeared earlier at several workshops \cite{isik2023information, lecomte2023information, schaeffer2023information1, schaeffer2023information2}.


\bibliographystyle{abbrvnat}
\bibliography{references_rylan}

\clearpage
\appendix
\onecolumn
\section{Proof of Theorem~\ref{theorem:recon_plus_uniform_min_mmcr}}
\label{sec:proof_nuclear_norm_lower_bound}

Recall that $\mathcal{L}_{MMCR} = - \|C\|_*$ is minimized when $\|C\|_*$ is maximized and that $\|C\|_*$ is upper bounded by $\sqrt{ND}$ if $N> D$ and $N$ if $N < D$ (Proposition~\ref{prop:nuclear_norm_upper_bound}). We want to show a network that achieves perfect reconstruction and perfect uniformity achieves this upper bound on the nuclear norm (equivalently, lower bound on the MMCR loss).

Following the proof of Proposition~\ref{prop:nuclear_norm_upper_bound}, let $\sigma_1, \ldots, \sigma_{\min(N,D)}$ denote the singular values of $C$, so that $\|C\|_* = \sum_i \sigma_i$. By Proposition~\ref{prop:means_on_sphere}, we have
$$\sum_i \sigma_i^2 = \Tr\!\left[C C^T\right] = \sum_{n=1}^{N}\bm\mu_n^T \bm \mu_n = N.$$
Now, by the equality version of Cauchy--Schwarz on the sequences $(1, \ldots, 1)$ and $\left(\sigma_1, \ldots, \sigma_{\min(N,D)}\right)$, we have
\begin{equation}\label{eq:equality-cauchy-schwarz}
\sum_i \sigma_i = \sqrt{\min(N,D)\left(\sum_i \sigma_i^2 - \sum_i\left(\sigma_i - \frac{\sum_j \sigma_j}{\min(N,D)}\right)^2\right)}.
\end{equation}
So if we can bound this ``variance'' of the singular values $\sum_i\left(\sigma_i - \frac{\sum_j \sigma_j}{\min(N,D)}\right)^2$, we can show that $\|C\|_*$ closely matches the upper bound obtained in Proposition~\ref{prop:nuclear_norm_upper_bound}.

To do this, let us consider matrix $\sqrt{D}C$. The vectors $\bm\mu_n$ are uniform over the $D$-dimensional hypersphere $\mathbb{S}^{D}$, so its rows $\sqrt{D}\bm\mu_n$ have mean zero, are isotropic, and (by Example~5.25 in Vershynin (2012) \citep{vershynin2012compressed}) are sub-gaussian with parameter $\|\sqrt{D}\bm\mu_n\|_{\psi_2} = O(1)$.\footnote{Here, $\|\cdot\|_{\psi_2}$ denotes the sub-gaussian norm (intuitively, the ``effective standard deviation'' of a sub-gaussian random variable). For a scalar random variable $X$, it is defined as $\|X\|_{\psi_2} \coloneqq \sup_{p \ge 1}p^{-1/2}(\mathbb{E}[|X|^p])^{1/p}$ (Definition~5.7 in Vershynin (2012)\citep{vershynin2012compressed}), and for a random vector $\bm{u} \in \mathbb{R}^D$, it is defined as $\|\bm{u}\|_{\psi_2} \coloneqq \sup_{\bm{v} \in \mathbb{S}^D}\|\bm{u}^T \bm{v}\|_{\psi_2}$ (Definition~5.22 in~\cite{vershynin2012compressed}).} Therefore,
\begin{itemize}
\item \textbf{If} $\mathbf{N \le D}$, then (using the fact that $\|\bm\mu_n\|_2=1$ for all $n \in [N]$) we can to apply Theorem~5.58 in Vershynin (2012)\citep{vershynin2012compressed} on the transpose of $\sqrt{D}C$, obtaining that for any $t \ge 0$, the singular values of $\sqrt{D}C$ are within $\sqrt{D} \pm O(\sqrt{N}) + t$ with probability at least $1-2\exp(-\Omega(t^2))$. Setting $t$ to a large enough multiple of $\sqrt{N}$, they are all within $\sqrt{D} \pm O(\sqrt{N})$ with probability at least $1-2 \exp(-N)$. Consequently, with the same probability, the singular values of $C$ are all within $\pm O(\sqrt{N/D})$ of each other, and we get $\sum_i\left(\sigma_i - \frac{\sum_j \sigma_j}{\min(N,D)}\right)^2 \le N \cdot O\left(\sqrt{N/D}\right)^2 = O(N^2/D)$. Plugging this into Eqn. \ref{eq:equality-cauchy-schwarz}, we get $\|C\|_* \le \sqrt{N (N - O(N^2/D))} = \sqrt{N}(1-O(N/D))$.
\item \textbf{If} $\mathbf{N \ge D}$, then we can apply Theorem~5.39 in Vershynin (2012) \citep{vershynin2012compressed} on $\sqrt{D}C$, obtaining that for any $t \ge 0$, the singular values of $\sqrt{D}C$ are within $\sqrt{N} \pm O(\sqrt{D}) + t$ with probability at least $1-2\exp(-\Omega(t^2))$. Setting $t$ to a large enough multiple of $\sqrt{D}$, they are all within $\sqrt{N} \pm O(\sqrt{D})$ with probability at least $1-2 \exp(-D)$. Consequently, with the same probability, the singular values of $C$ are all within $\pm O(1)$ of each other, and we get $\sum_i\left(\sigma_i - \frac{\sum_j \sigma_j}{\min(N,D)}\right)^2 \le D \cdot O(1)^2 = O(D)$. Plugging this into Eqn. \ref{eq:equality-cauchy-schwarz}, we get $\|C\|_* \le \sqrt{D (N - O(D))} = \sqrt{ND}(1-O(D/N))$.
\end{itemize}

\clearpage
\section{Python Code for Perfect Reconstruction and Perfect Uniformity Embeddings}
\label{app:python_code}

To test our claim that networks which achieve perfect reconstruction and perfect uniformity achieve the nuclear norm upper bound, we sample a uniform distribution of centroids (thereby enforcing reconstruction by construction) and measure the nuclear norm relative to our claimed upper bound. Python code for our simulations is included below:

\begin{lstlisting}[language=Python]
import pandas as pd
import numpy as np


N_list = np.logspace(start=1, stop=4, num=11).astype(int)
D_list = np.logspace(start=1, stop=4, num=11).astype(int)
repeats = np.arange(5).astype(int)
uniform_distribution_nuclear_norm_data_list = []

for N in N_list:
    for D in D_list:
        print(f"N: {N}\tD: {D}")
        for repeat in repeats:
            embeddings = np.random.normal(loc=0, scale=10.0, size=(N, D))
            embeddings /= np.linalg.norm(embeddings, axis=1, keepdims=True)
            row = {
                "Spectrum": "uniform",
                "Number of Data Manifolds (N)": N,
                "Embedding Dimensionality (D)": D,
                "Repeat": repeat,
                "Nuclear Norm": np.linalg.norm(embeddings, ord="nuc"),
            }
            uniform_distribution_nuclear_norm_data_list.append(row)

uniform_distribution_nuclear_norm_df = pd.DataFrame(
    uniform_distribution_nuclear_norm_data_list
)
\end{lstlisting}

\section{Multimodal Maximum Manifold Capacity Representations}
\begin{figure*}[t]
    \centering
    \includegraphics[width=0.49\textwidth]{figures/image_text/imagenet_zeroshot_top1_vs_epoch_lmplot.png}%
    \includegraphics[width=0.49\textwidth]{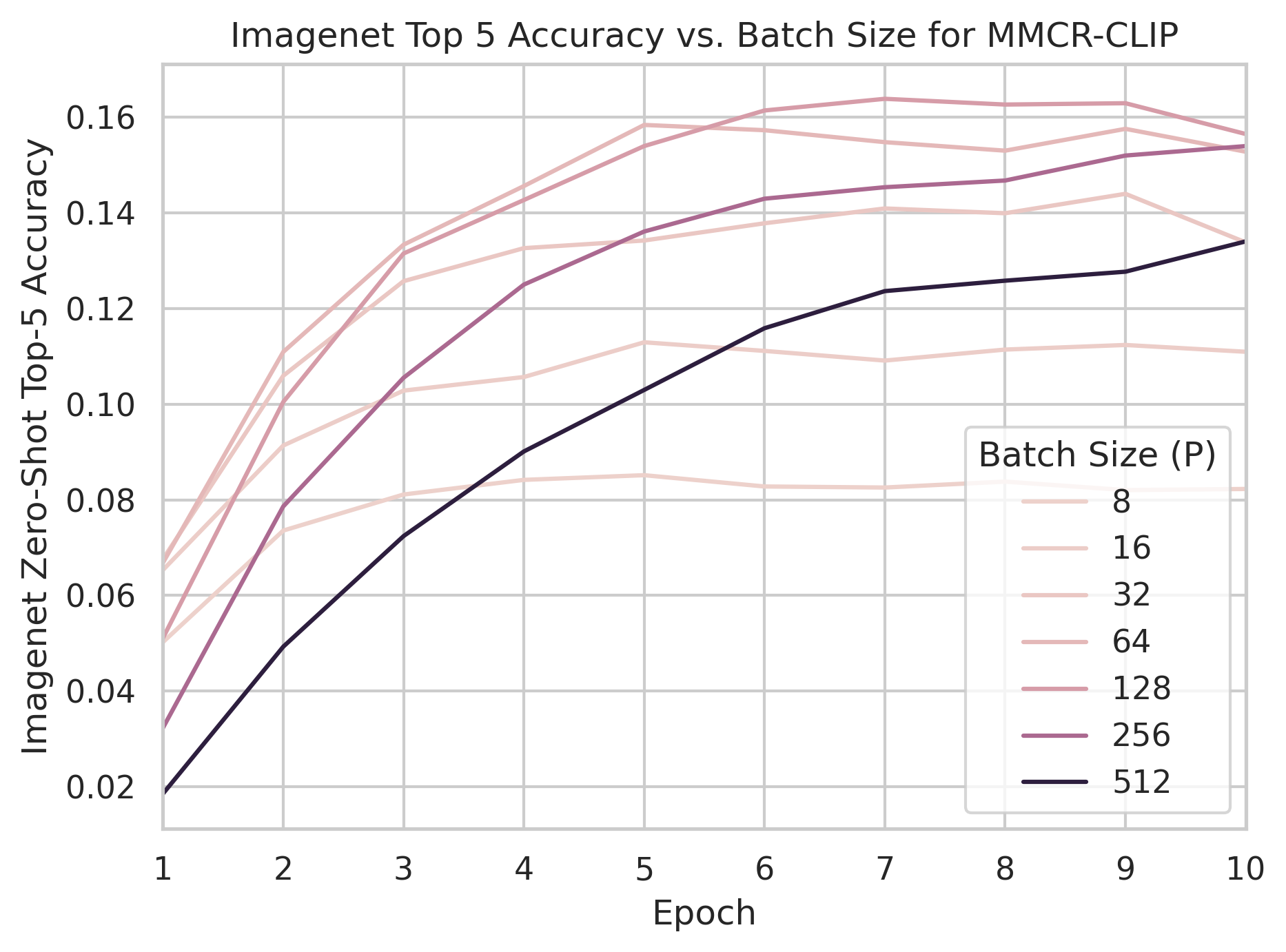}
    \caption{Multimodal MMCR exhibits strong batch size dependence in ImageNet zero-shot validation performance. Intermediate batch sizes, such as $128$ and $256$, achieved the best validation performance by a large margin. By contrast, the smallest batch sizes or batch sizes closest to the embedding dimension size of $1024$ fared the poorest. }
    \label{fig:mmcr_clip_val}
\end{figure*}


\begin{figure*}[t]
    \centering
    \includegraphics[width=0.5\textwidth]{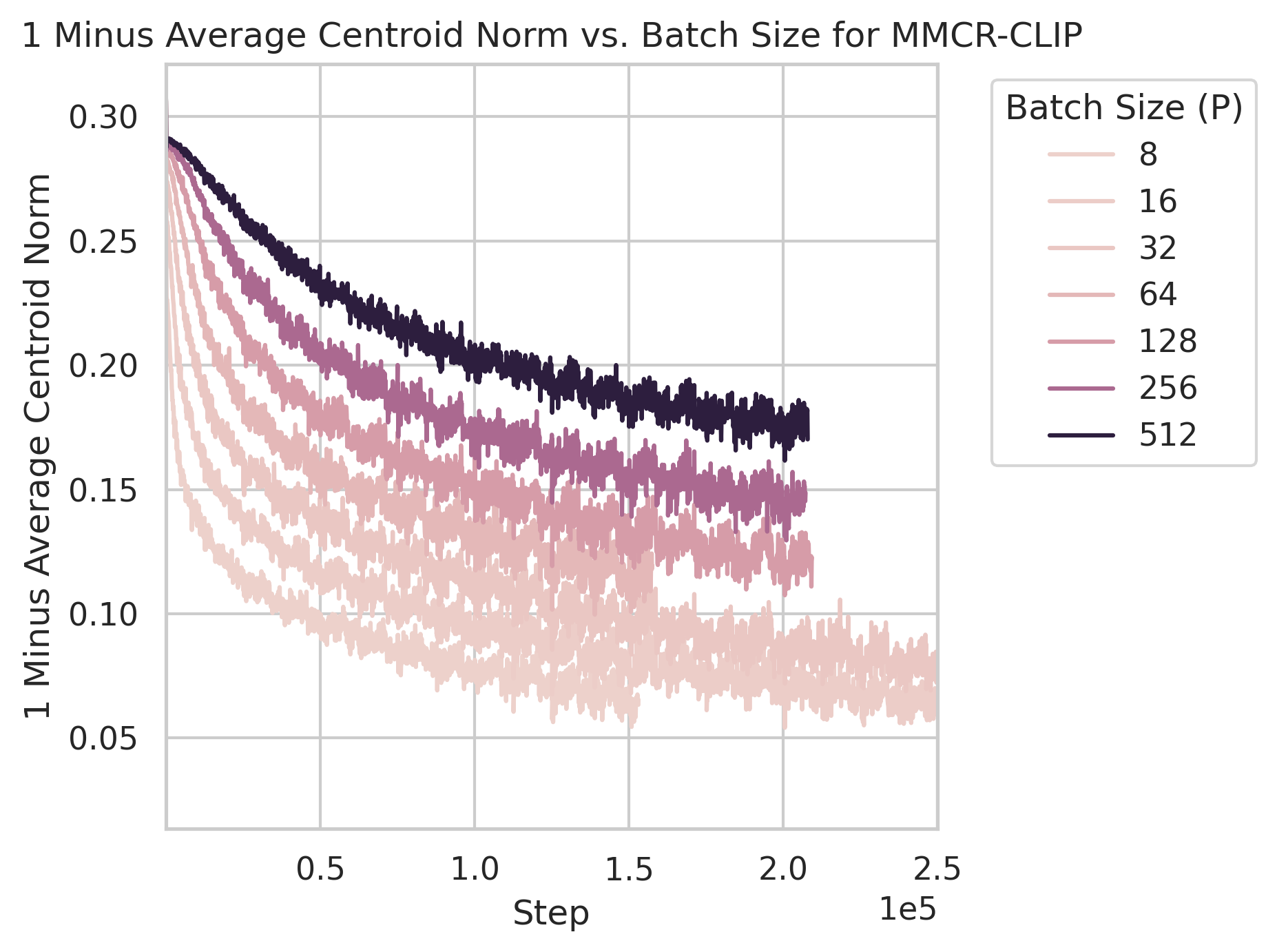}%
    \includegraphics[width=0.49\textwidth]{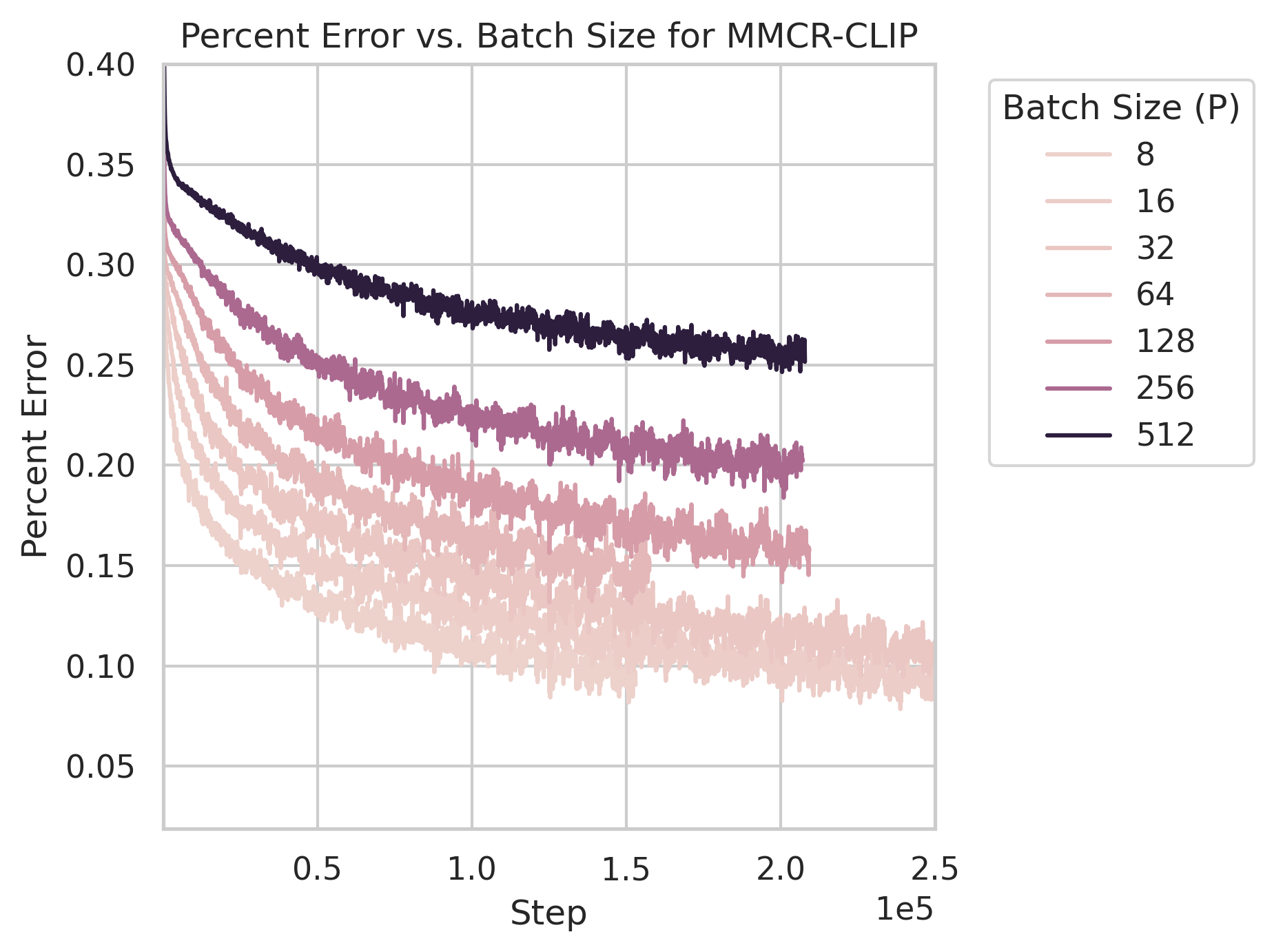}
    \caption{To understand the perplexing batch size dependence, we analyze the complement of the average centroid norm ($1-\lVert \mu \rVert_2^2$) and the pretaining percent error relative to the lower bound as defined earlier($1-\frac{\lVert M \rVert_*}{N}$). The complement of the average centroid norm is an unbiased estimator for perfect reconstruction in our network. We find that lower batch sizes converge closer to perfect reconstruction and to lower percent error. } 
    \label{fig:mmcr_clip_train}
\end{figure*}

\end{document}